**Digitising Cultural Complexity: Representing Rich Cultural Data in a Big Data environment.**

Jennifer Edmond (edmondj@tcd.ie) & Georgina Nugent Folan (nugentfg@tcd.ie)
Trinity College Dublin, Dublin, Ireland

**Introduction**

One of the major terminological forces driving ICT integration in research today is that of "big data." While the phrase sounds inclusive and integrative, "big data" approaches are highly selective, excluding input that cannot be effectively structured, represented, or digitised. Data of this complex sort is precisely the kind that human activity produces, but the technological imperative to enhance signal through the reduction of noise does not accommodate this richness.

Data and the computational approaches that facilitate "big data" have acquired a perceived objectivity that belies their curated, malleable, reactive, and performative nature. In an input environment where *anything* can "be data" once it is entered into the system as "data," data cleaning and processing, together with the metadata and information architectures that structure and facilitate our cultural archives, acquire a capacity to delimit what data are. This engenders a process of simplification that has major implications for the potential of future innovation within research environments that depend on rich material yet are increasingly mediated by digital technologies. It is the objective of the KPLEX project, lead by Trinity College Dublin, to explore the impact of bias in big data approaches to knowledge creation, by investigating the delimiting effect digital mediation and datafication has on rich, complex cultural data.

The issues of concern to this project are all of relevance to our ways of being in the digital age, but for the purposes of this presentation, we will focus on three of the themes that the larger KPLEX project is teasing out, namely:

> i) How the word data is used, or indeed misused, in scientific discourse and the implications these discursive practices have.

> ii) How our strategies for navigating the data deluge, via metadata, keywords, search and other algorithmic or organisational strategies can restrict the interpretative potential of data, and make us susceptible to an implicit truth claim made not on the basis of careful consideration or deep faith, but on algorithmic or digital bias.

> iii) How approaches to knowledge creation found in the arts and humanities can help reconceptualise the ways we speak about and use data, both as citizens and as scholars.

**Section 1. Data**

I'm going to begin by reviewing some of the existing implicit definitions of data that underlie ICT-driven research, before moving on to present some of our preliminary findings; specifically in relation to the usage of the term "data" in computer science journal articles, and the attitudes towards data that underlie computer science research and development.

In "Data before the Fact," Daniel Rosenberg outlines the history of "data" as a concept prior to the 20th century, exploring how it acquired its "pre-analytical, pre-factual status."[1] Jesse Rosenthal similarly presents data as an entity that "resists analysis," an "object that cannot be questioned."[2] Christine Borgman further elaborates on this, stating that "Data are neither truth nor reality."[3] It's fitting then, that Rosenberg famously cites data as "rhetorical"[4] while Rita Raley argues data is "performative,"[5] and perhaps these performative and rhetorical facilities explain data's mutative nature.

Our research has found that inconsistent and contradictory statements on data are multifarious. While this is perhaps to be expected among the more philosophical or conceptual approaches to defining data I've just covered, inconsistencies and tacit disagreements over what data are also exist within the computer science community, though this huge variability goes largely unsurfaced and undiscussed in the scientific literature: indeed, it seems to be an acceptable part of the discourse. In a single article,[6] the word data can acceptably be used to refer to something newly drawn out of the environment ('captured', 'collected') or as something already available for access, analysis, navigation. It can refer to complex hybrid objects ('experimental data,' 'digital data'), or comparatively simply strings of regular meter readings ('sensor data'). In practice, as in theory, data is both wild and tamed, simple and complex, pre-epistemic and pre-processed, of direct use to humans and purely machine readable. Of greatest interest (and concern) it also seems to accrete various, largely positive qualities, according to which it may be 'relevant,' 'contextual,' 'various,' 'external,' 'complex' or 'rich,' etc.

The scale of this polysemy is quite striking and leads as it does to an astonishing reliance on the single term 'data' within the scientific discourse. Searching for occurrences of the terms "data" and "big data" in five articles selected at random from the *Journal of Big Data* 2016, results in over 650 occurrences of the term data, including 325 in the composite phrase 'big data' and 124 in the context of a 'dataset' or 'database.'

---

[1] Rosenberg, "Data before the Fact," in Gitelman, "Raw Data" is an Oxymoron, 18.
[2] Jesse Rosenthal, "Introduction: 'Narrative against Data,'" *Genre* 50, no. 1 (April 1, 2017): 1., doi:10.1215/00166928-3761312.
[3] Christine L. Borgman, "Big Data, Little Data, No Data," MIT Press, 17, accessed April 7, 2017, https://mitpress.mit.edu/big-data-little-data-no-data.
[4] Rosenberg, "Data before the Fact," in Gitelman, *"Raw Data" Is an Oxymoron*, 18.
[5] Rita Raley, "Dataveillance and Countervailance" Gitelman, *"Raw Data" Is an Oxymoron*, 128.
[6] Tabard A, Hancapié-Ramos JD, Esbensen M, and Bardram J (2011) The eLabBench: An Interactive Tabletop System for the Biology Laboratory. *ITS 11.*

| *Journal of Big Data*, 2016 | | | |
|---|---|---|---|
| Article title: | Total occurrences of the term "data." | Total occurrences of the term "big data." | Total occurrences of the term "database"/ "dataset." |
| Khalilian et al., J Big Data (2016) 3:1. | 234 | 4 | 62 |
| Bughin, J Big Data (2016) 3:14. | 35 | 176 | 2 |
| Kintsakis et al., J Big Data (2016) 3:20. | 128 | 9 | 125 |
| Suthakar et al., J Big Data (2016) 3:21. | 200 | 19 | 19 |
| van Altena et al., *J Big Data* (2016) 3:23. | 62 | 113 | 16 |
| Totals combined: | 659 | 325 | 124 |

If you think this is bad, by the way, a 2015 article by Najafabadi et al.,[7] sees the term data used an incredible 507 times over 21 pages (24.14 times/page). Of these, 124 iterations pertain to "big data" which leaves 383 uses of "data" proper.)

Within this corpus we observe such inconsistencies as idiosyncracies of spelling key terms within the one paper (dataset / data set); the same term used to refer to specific data/ datasets & more general data / datasets; the same term used to refer to different phases of investigation; to different **data** (stream/ cluster/ original / evolved/ evolving), and to data as a synecdochal term covering both the whole and the part. The following quotations give a sense of what one finds:

> "***Data pretreatment*** *module is outside from online component and it is done to **preprocess stream data** from the **original data** which is produced by the previous component in the form of **data stream**.*"

---

[7] Najafabadi et al., "Deep learning applications and challenges in big data analytics," Journal of Big Data (2015), 2:1

*"In addition, we calculate the standard deviation for the entire data in the stream to check whether all the data are of the same value or not."*

*"Due to visiting data once during the processing data in stream, the performance of processing data is crucial."*[8]

One fares no better with the phrase 'big data.' Although our findings remain based on a relatively small sample set so far, not only does the phrase remain highly malleable, but it seems to have been infused with a corporate glow of being the cure for just about any ill,[9] as implied by the image below.

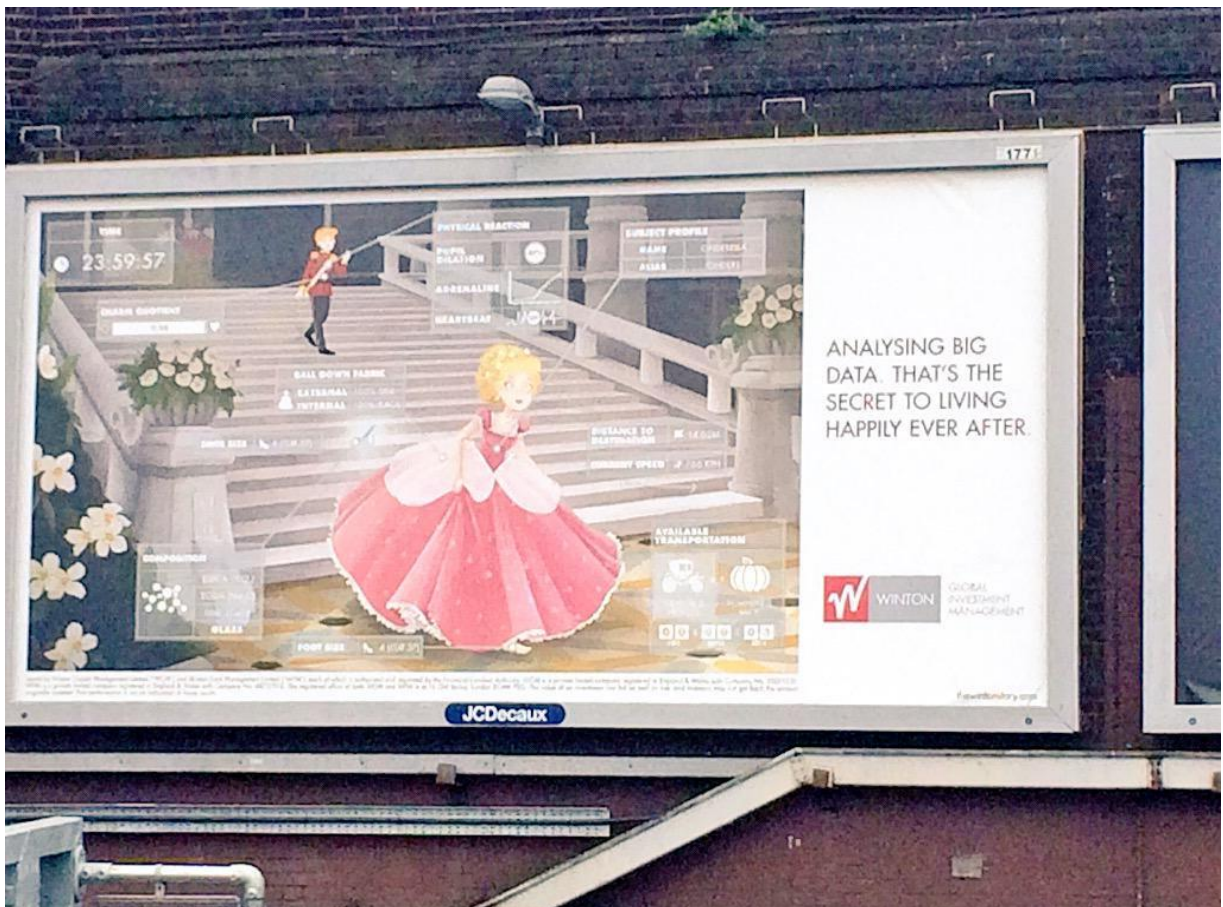

Image copyright Winton Global Investment Management, 2015.

Problems identifiable at the level of "data" or singular "data points" are only magnified when we take it to the scale of big data. To paraphrase, big data magnifies these problems bigly.

In addition to a lack of transparency in relation to the transformations that are applied to data, the technique of cleaning data so as to produce more refined data is not systematised. Rather, it is presented as an almost esoteric process and one that brings with it implicit assumptions regarding the methods used to clean or process; assumptions that are not only discipline specific, but often researcher specific. The fact that salient aspects of an original record might have been lost along the way as data passes through the hands and servers of different actors is not only not recognised as potentially problematic, it is recognised as a part of the scientific process. The NASA protocol for describing the processing levels of research data, for example, carefully grades data from level 0 (raw or native data) through 4 additional levels of processing. However, these distinctions only pertain to the onset of the research, the point where data is gathered; thereafter the number resets and the data is considered raw, until it is subjected to further processing.

---

[8] Khalilian et al., "Data stream clustering by divide and conquer approach based on vector model,," *J Big Data* (2016) 3:1.
[9] Bughin, J Big Data (2016) 3:14.

One might be inclined to see it as typical humanist hairsplitting to point out that our fellow researchers from across the epistemic divide use words less rigorously than we might like. But we would contend that these discursive practices carry a high social and scientific price. In a recent Guardian article about the data held by Tinder, Paul-Olivier Dehaye from the organisation personaldata.io stated "We are leaning towards a more and more opaque society, towards an even more intangible world where data collected about you will decide even larger facets of your life. Eventually, your whole existence will be affected."[10] We can't feel our data bodies—our "data doubles"[11]—, or hear the symphonies we may be performing with the digital instruments that are so much a part of our daily lives. Data has become a modern pharmakon, both restorative and deadly. How much of this is due not only to the immateriality of data, but also to the poverty of our vocabulary for speaking about it, conflating inputs, outputs and middleware states, rights implicit and rights granted, spanning a huge variety of practices. A computer scientist's lack of sensitivity to input data and output data, to the raw and the cooked, so to speak, may seem innocuous, until you realise the values underlying these practices are passed directly to the next generation. At this point the non-distinction between input and output becomes a lack of distinction between 'my data' and 'your data,' and this has emerged as one of the key recurring themes in the detailed interviews we conducted with computer scientists on data, interviews we are currently in the process of transcribing and encoding. A lack of concern for how data changes as it is removed from its provenance and transformed, becomes a lack of alarm when two data sets become combined to enable an individual's identity to be determined and this knowledge abused. Words matter, particularly when they directly relate to our ways of being in a digital age, and no one should be held to a higher standard in this than scientists.

**Section 2: Metadata**

We know, or have known from at least as early as 1951 and Suzanne Briet's *Qu'est-ce que la documentation*? that just like Briet's archival document, "Data are not pure or natural objects with an essence of their own."[12] We also know that data "exist in a context, taking on meaning from that context and from the perspective of the beholder. "[13] But questions remain as to how much of this context is being redacted, modified or hidden by the surrounding information architecture. In the transition from analogue to digital, data loses facets of its native context, and acquires new contexts in the form of the metadata that situate it in the digital environment. To what extent is so-called "raw" or "native" data shaped by the database and contexts imported onto it by metadata in and of itself? And what of the contexts that accompanied the data within its "original," "proto-data," or "native" environment, its native contexts, how are these transferred to and catered for in the database?

---

[10] Paul-Olivier Dehaye, Quoted in Judith Duportail, "I asked Tinder for my data. It sent me 800 pages of my deepest, darkest secrets," *The Guardian*, 26/09/17. Accessed: 27/09/17.
[11] Ericson and Haggerty, 2006; Raley, 2013.
[12] Borgman, "Big Data, Little Data, No Data," 18.
[13] Borgman, "Big Data, Little Data, No Data," 18.

      Classification systems and documentation standards are based upon high level, non granular categories. Such categories are an advantage because they allow for the classification of diverse and abundant quantity of data, but a disadvantage because they flattens out the data and makes the granular more difficult to identify and access. In the previous section I introduced Raley's concept of data as "performative."[14] Others refer to how data can become "a sort of actor" "reshaping"[15] the social world or how they carry the imprint of "the framework through which they were constructed, collected or collated."[16] Now if data are performative and possesses the capacity to shape and be shaped by its contexts, both native and acquired, then so is metadata. This effect is only magnified when we are working on the scale of big data, and is particularly problematic when we are dealing with "uncertain," "ambiguous" or "complex" data. The performative affect of metadata—intended, deliberate, or unforeseen—is a phenomenon well illustrated by Todd Presner's analysis of the relationship between research subject, research data, and system metadata in the Shoah Visual History Archive.

      It goes without saying that great sensitivity is required of any project dealing with the personal testimonies of victims of the holocaust. It is far too difficult to err either on the side of losing sight of the many many victims who died anonymously (by focussing on the more famous stories like Elie Wiesel or Anne Frank), or to lose sight of the human suffering at the core of the holocaust experience by focussing on the scale of the tragedy (a many have said of the Digital Monument to the Jewish Community in the Netherlands). An algorithmic approach to the records of the holocaust should, in theory, be able to reconcile these poles of human frailty in the face of enormous data corpora. However, as Presner observes, the Shoah VHA's was reliant on human cataloguers for the creation of metadata and finding aids, leading to an approach to uncertainty within the digital archive that saw certain material categorised as "'indeterminate data' such as 'non-indexable content.'"[17] Presner is pragmatic in observing that "databases can only accommodate unambiguous enumeration, clear attributes, and definitive data values; *everything else is not in the database*."[18] But that which is considered ambiguous differs dramatically, ranging from repetition, pauses, emotion, noise or silence and even, most troublingly, material the indexer "doesn't want to draw attention to (such as racist sentiments against Hispanics, for example, in one testimony)."[19] What remains, as Presner notes, is "a

---

[14] Rita Raley, "Dataveillance and Countervailance" Gitelman, *"Raw Data" Is an Oxymoron*, 128.
[15] David Ribes and Steven J. Jackson, "Data Bite Man: The Work of Sustaining a Long-Term Study" ibid., 148.
[16] Drucker, 2014: 128.
[17] Presner, in Fogu, Claudio, Kansteiner, Wulf, and Presner, Todd, *Probing the Ethics of Holocaust Culture*, History Unlimited (Cambridge: Harvard University Press, 2015), http://www.hup.harvard.edu/catalog.php?isbn=9780674970519.
[18] Presner, in ibid.
[19] Presner, "The Ethics of the Algorithm: Close and Distant Listening to the Shoah Foundation Visual History Archive," *Probing the Ethics of Holocaust Culture*, History Unlimited (Cambridge: Harvard

kind of 'normative story' (purged of certain contingencies and unwanted elements)."[20]

We are interested in what happens to the "non-normative" sections of the dialogues or archival material, the material that is *not* assigned a keyword. Privileging one neglects and undercuts the importance of the other. Each section assigned a keyword has a relationship with the material on either side of it within the dialogue, but this material is left "silent," "hidden" (in certain cases, especially cases such as the one above where certain facets of the testimonies are "purged" (Presner's word) from the keyword thesaurus of the archive), or alternatively left latent and unstructured in the archive, left without a keyword. It is for this reason that traditionally trained archivists and librarians struggle with provenance standards such as that developed by the W3C: to even begin to represent the richness of an archival record, which may have been passed through numerous hands and families, or be marked with doodles or wine glass stains, or contain such a variety of material that its eventual relevance is hard to judge, transcends the kinds of clear categories and linear relationships such standards propose.

We should remember in this respect that metadata is not the only technique that hides from us as much as it possibly exposes. Leave aside the question of what role paid promotion plays in Google page rank algorithm, the fact remains that many of our historical records remain undigitised, and the provision of what we do have is highly uneven across regions and demographics. It is easy enough to create and promote alternative facts where the alternatives are also discoverable to the determined: it is easier still to deny history when the records lie locked up in archives that, to many, seem not to exist because they are resolutely analogue, and undiscoverable using contemporary methods.

**Section 3 : Practices (and conclusion)**

The investigations to date of the KPLEX project have exposed the extent to which computer scientists rely on the word "data," and some of the costs that society and science are incurring from this imprecision in an age when our data may be big, but our ability to express the heterogeneity within this is limited. To us, the answer lies in a closer investigation of the knowledge creation and management practices of the humanities. Humanists, actually resist the term data, and moreover, resist fully digital environments, recognising that unique place of the analogue, the material. In the place of the all-encompassing monolith of data (and its functional equivalents such as content, information, etc., which seem in the literature to be used with no real distinction) we encounter instead primary literature, secondary literature, theoretical and methodological texts; we have notes and annotations and drafts, we have digital finding aids and scholarly editions. Every one of these terms refers not only to a form of textual information, but also to their functions, their layers and level of epistemic pre-processing, their place in the current scholars' pursuit of knowledge, and that of others.[21]

For this reason, the nascent results of the KPLEX project propose the following measures by which greater richness and transparency can be brought to big data research through big data

---

University Press, 2015).
[20] Ibid.
[21] see https://hal.archives-ouvertes.fr/hal-01566290/document

discourse. We propose that data transformations should and must become a part of the provenance of a dataset, through a 'data passport.' We propose that researchers should have an instrument, similar to the ECDL, to monitor their development of the kinds of skills required to successfully work with composite, complex or personal data. We propose that consolidated digital environments are not enough - not for research and not for people. Developers often dislike the manner in which user input complicates technical workflows, but we need more of this, more adaptation to the embodied practices of knowledge creation that are hard wired into our kinetic processing. We need systems that expose their uncertainty in comprehensible ways for the average user, so we can see, for example, where machine translation strips away cultural complexity. And most importantly, we need the vocabulary and the knowledge to 'feel' data. As the author of the Guardian article quoted previously put it: *"we can't feel data. This is why seeing everything printed strikes you. We are physical creatures. We need materiality."*[22] Paradoxically, it may be here, in the the material, embodied and uncertain, that we will find the way forward for our ways of being in the digital age.

---

[22] Judith Duportail, "I asked Tinder for my data. It sent me 800 pages of my deepest, darkest secrets," *Guardian* 26/09/2017. http://www.theguardian.com/technology/2017/sep/26/tinder-personal-data-dating-app-messages-hacked-sold